\documentclass[conference]{IEEEtran}
\IEEEoverridecommandlockouts
\usepackage{amsmath,amssymb,amsfonts}
\usepackage{graphicx}
\usepackage{tikz}
\usepackage{pgfplots}
\usetikzlibrary{shapes,arrows.meta,positioning,fit,calc}
\usepackage{url}
\usepackage{cite}
\usepackage{float}
\usepackage{array}
\usepackage{booktabs}

\pgfplotsset{compat=1.18}
\begin{document}

\title{AI-Driven SaaS Platform Integration: AI Express for Unified Generative AI Services with Pay-As-You-Go Credit Systems and Scalable Microservice Architecture}

\author{
\IEEEauthorblockN{Deepit Sapru}
\IEEEauthorblockA{University of Illinois Urbana-Champaign}
\IEEEauthorblockA{Email: dsapru2@illinois.edu}
}

\maketitle

\begin{abstract}
Generative AI is currently fragmented, with separate subscriptions required for text generation, image synthesis, and code assistance, leading to high costs and workflow disruptions. AI Express is a SaaS platform that consolidates these capabilities into a unified pay-as-you-go service. The platform provides a range of AI tools—including a Core Code Reviewer that evaluates GitHub repositories, a Text-to-Website generator that converts text into fully functional websites, script generation, grammar checking, data analysis, and watermark removal—all accessible through a single interface without requiring technical expertise.
\end{abstract}


\section{Introduction}

The proliferation of generative artificial intelligence has substantially transformed digital productivity, software engineering practices, and media generation paradigms. Nevertheless, the contemporary ecosystem of generative AI remains remarkably fragmented, a condition exacerbated by steep operational expenditures \cite{Maddukuri2025}. Professionals, researchers, and students frequently find themselves compelled to maintain concurrent subscriptions to multiple premium services merely to access a diverse array of text generation, image synthesis, and code assistance capabilities. This disjointed arrangement not only inflates costs but also imposes significant cognitive overhead arising from context switching among distinct interfaces and authentication systems \cite{Bhargava2025}.

In response to these inefficiencies, we introduce AI Express, a Software-as-a-Service (SaaS) platform engineered to aggregate state-of-the-art AI models within a unified, cohesive, and user-centric environment. The system functions as a multi-modal hub, integrating both locally hosted and externally invoked AI engines to deliver a comprehensive toolkit. The core offerings encompass: (i) Core Code Reviewer, an automated static analysis tool that scrutinizes GitHub repositories for code quality, security vulnerabilities, and adherence to best practices; (ii) AI Data Analyst, a structured data processing module capable of ingesting CSV, TSV, or Excel files and responding to natural-language analytical queries; (iii) content development utilities, including a Script Generator that produces structured documents from brief semantic descriptions and a Grammar Checker that refines human-authored prose; (iv) digital image processing capabilities, particularly a watermark removal tool based on deep learning inpainting; and (v) extensible auxiliary services such as an AI Web Scraper and an AI Resume Builder.

The architectural backbone of AI Express adopts a microservices paradigm, implemented with React.js for the frontend, Node.js for the backend orchestration, and PostgreSQL for persistent data storage. This design ensures horizontal scalability and operational resilience under varying load conditions, following principles highlighted by \cite{Lakkarasu2025}. A pivotal innovation of the platform is its pay-as-you-go credit economy, which supplants conventional fixed monthly subscriptions. This mechanism enables users to purchase credits that are consumed proportionally to the computational intensity of each service invocation, thereby rendering premium AI capabilities accessible to budget-constrained students and independent developers \cite{Seuss2025}.

This paper presents a comprehensive exposition of the AI Express framework, delineating its architectural design, implementation details, performance characteristics, and comparative advantages over existing solutions. The remainder of the document is structured as follows: Section II reviews related work; Section III details the system architecture and design methodology; Section IV describes implementation and module functionalities; Section V presents performance evaluation and comparative analysis; and Section VI concludes with future directions.

\section{Related Work}

Recent work on AI-driven SaaS platforms has focused on integrating multiple intelligent services. \cite{Neursu2025} analyzed how AI transforms customer experience optimization in SaaS products, while \cite{Kalathot2025} explored dynamic user experience personalization in similar environments. The challenge of unifying diverse AI tools under a single economic model remains, as many solutions still impose separate subscriptions.

Large language models (LLMs) have been applied to code review and data analytics. \cite{Dhameliya2026} demonstrated that LLM-assisted code inspection significantly improves bug detection, but such tools are often offered as isolated plugins. Similarly, \cite{Arora2024} showed that AI-driven analytics can enhance business intelligence in SaaS, yet these capabilities are rarely bundled with code and image services.

In image processing, deep learning inpainting methods have achieved high fidelity in watermark removal. \cite{Nagamani2026} accelerated unsupervised feature learning for denoising autoencoders, providing a foundation for real-time artifact removal. However, these computationally intensive tasks are typically confined to desktop applications.

Microservice architectures offer a scalable way to orchestrate multimodal AI workflows. \cite{Vanama2025} presented AI-driven cloud integration for enterprise systems, and \cite{Naveen2024} surveyed cloud computing for data-driven AI services. The pay-as-you-go credit model adopted here differentiates AI Express from subscription-based platforms, directly addressing the cost barrier faced by students and independent developers \cite{Beesabathuni2026}.

\section{System Architecture and Design Methodology}

The client-side user interface is implemented as a single-page React.js application written in TypeScript. TypeScript ensures type-safety for dynamically creating and updating components. We use Redux for managing state and Material-UI for design language consistency. Our front end captures user inputs; in the case of the grammatical error checker it will be a paragraph of text, and for the watermark remover it will be the binary data of an image. It also completes client-side validation \cite{Kalathot2025}.

The request that comes in goes through a layer of security and routing. At this point, the request is routed to the appropriate microservice after performing the required JWT authentication, request validation, and rate limiting. In addition, NGINX works as a reverse proxy that is responsible for SSL termination and load balancing requests to multiple Node.js worker processes. This enables the system to scale horizontally with user demand \cite{PatelA2026}.

The backend environment consists of a collection of independent service nodes. Each of these hosts AI models responsible for a certain AI domain.
\begin{itemize}
\item The first node connects networks for natural language. The tools encompass the Script Generator, Grammar Checker, and static code analysis using AST for the Code Reviewer.
\item The second node hosts deep learning digital inpainting, employing temporal convolutional networks (TCN) based algorithms to perform watermark removal.
\item The third node examines files with structured data such as CSV, TSV, and Excel sheets, and produces an executable script in SQL or Python according to user requests.
\end{itemize}

Data Persistence Layer utilizes PostgreSQL, a well-known relational database. Redis helps in caching sessions and also acts as a rate limiter \cite{Sarakam2026}.

Data structure includes tables for users (login credentials, profile info, balance credits, and usage history), pricing tables for available AI services (computational weight), tables that store transactions (debits in credits with timestamps), and tables for content (uploaded datasets and generated content).

Security expertise begins at the original design, embedding encryption at every layer. Passwords are hashed using bcrypt with 12 salt rounds. JWT tokens are designed to timeout after a 24-hour period with refresh token rotation. Inputs are validated properly to prevent SQL injection and cross-site scripting attacks. Uploads of at most 10 MB in size are validated by MIME type and scanned for viruses \cite{Battapothu2023}.

The uniqueness of our architecture lies in the credit-based economy that resolves the economic problem faced by subscription models. Each microservice has a computational weight which determines the computational resources used. A user incurs the credit cost when a task finishes \cite{Beesabathuni2026}.

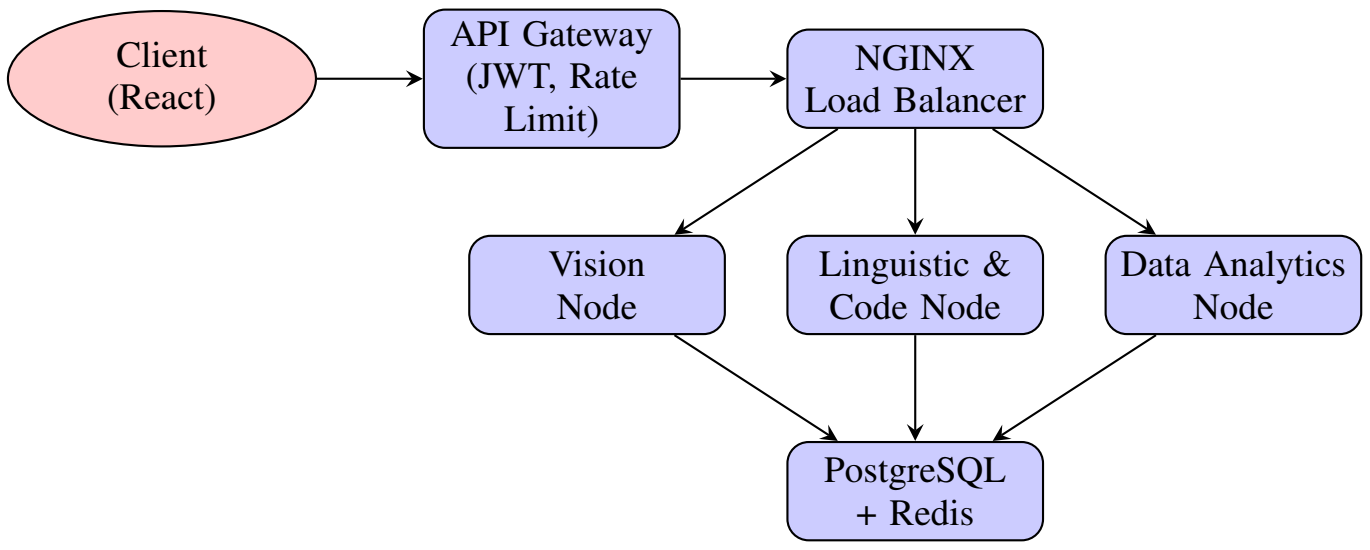
\begin{figure}[!t]
\centering
\begin{tikzpicture}[
    node distance=0.55cm and 0.65cm,
    every node/.style={font=\scriptsize},
    block/.style={
        rectangle,
        draw,
        fill=blue!20,
        text width=1.55cm,
        text centered,
        minimum height=0.65cm,
        inner sep=2pt,
        rounded corners
    },
    cloud/.style={
        ellipse,
        draw,
        fill=red!20,
        text width=1.3cm,
        text centered,
        minimum height=0.55cm,
        inner sep=2pt
    },
    line/.style={
        draw,
        ->,
        >=stealth
    }
]

% Client and infrastructure
\node[cloud] (client) {Client\\(React)};
\node[block, right=0.7cm of client] (gateway) 
    {API Gateway\\(JWT, Rate Limit)};
\node[block, right=0.7cm of gateway] (router) 
    {NGINX\\Load Balancer};

% Microservice nodes
\node[block, below=0.7cm of router] (ling) 
    {Linguistic \&\\Code Node};
\node[block, left=0.4cm of ling] (vision) 
    {Vision\\Node};
\node[block, right=0.4cm of ling] (data) 
    {Data Analytics\\Node};

% Database
\node[block, below=0.7cm of ling] (db) 
    {PostgreSQL\\+ Redis};

% Connections
\path[line] (client) -- (gateway);
\path[line] (gateway) -- (router);

\path[line] (router) -- (ling);
\path[line] (router) -- (vision);
\path[line] (router) -- (data);

\path[line] (ling) -- (db);
\path[line] (vision) -- (db);
\path[line] (data) -- (db);

\end{tikzpicture}
\caption{System architecture of AI Express illustrating the client, API gateway, load balancer, microservice nodes, and persistent storage.}
\label{fig:arch}
\end{figure}

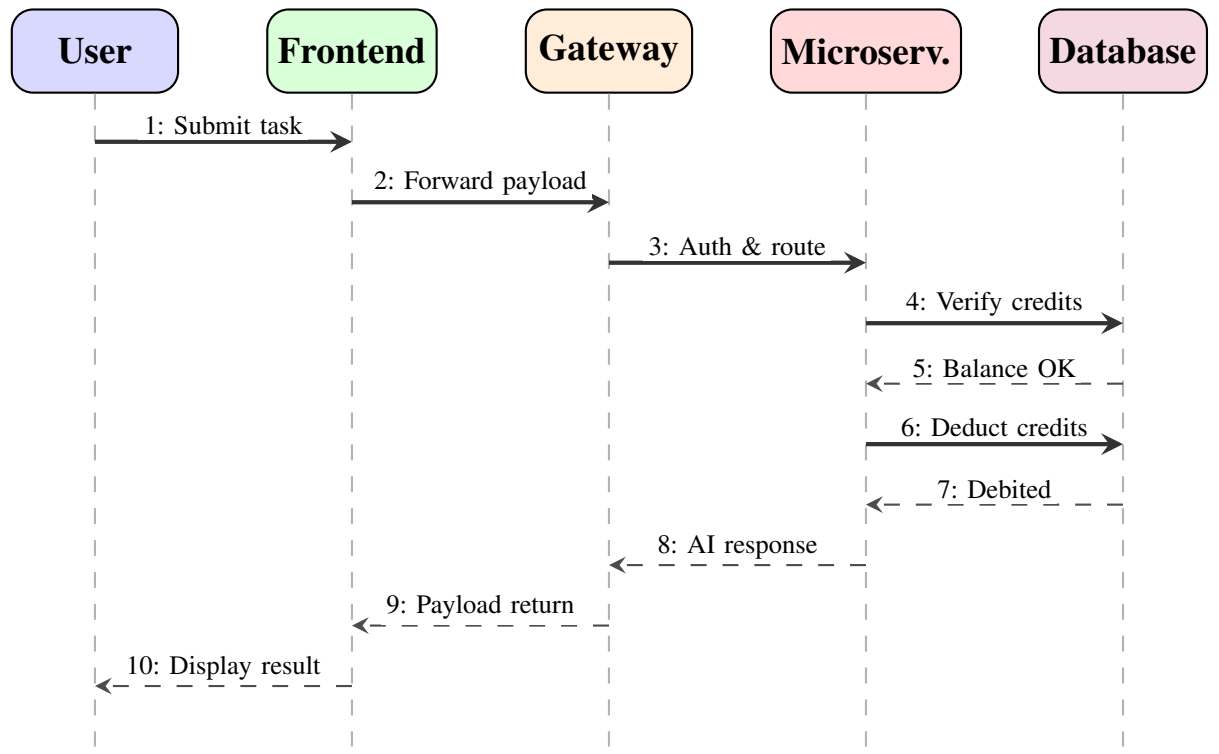
\begin{figure}[!t]
\centering
\begin{tikzpicture}[
    every node/.style={font=\scriptsize},
    entity/.style={
        rectangle,
        draw,
        minimum width=1.1cm,
        minimum height=0.55cm,
        inner sep=2pt,
        rounded corners,
        font=\scriptsize\bfseries
    },
    msg/.style={
        ->,
        >=stealth,
        draw=black!80,
        thick
    },
    ret/.style={
        ->,
        >=stealth,
        dashed,
        draw=black!70
    },
    lbl/.style={
        fill=white,
        inner sep=1pt,
        font=\tiny
    }
]

% Lifeline Headers
\node[entity, fill=blue!15]   (user)  at (0,0)    {User};
\node[entity, fill=green!15]  (front) at (1.7,0)  {Frontend};
\node[entity, fill=orange!15] (gw)    at (3.4,0)  {Gateway};
\node[entity, fill=red!15]    (ms)    at (5.1,0)  {Microserv.};
\node[entity, fill=purple!15] (db)    at (6.8,0)  {Database};

% Lifelines (vertical dashed lines)
\draw[dashed, gray!60] (user)  -- (0,-4.6);
\draw[dashed, gray!60] (front) -- (1.7,-4.6);
\draw[dashed, gray!60] (gw)    -- (3.4,-4.6);
\draw[dashed, gray!60] (ms)    -- (5.1,-4.6);
\draw[dashed, gray!60] (db)    -- (6.8,-4.6);

% Message sequence (top to bottom)
\draw[msg] (0,-0.6) -- node[lbl, above] {1: Submit task} (1.7,-0.6);
\draw[msg] (1.7,-1.0) -- node[lbl, above] {2: Forward payload} (3.4,-1.0);
\draw[msg] (3.4,-1.4) -- node[lbl, above] {3: Auth \& route} (5.1,-1.4);

% Credit Check
\draw[msg] (5.1,-1.8) -- node[lbl, above] {4: Verify credits} (6.8,-1.8);
\draw[ret] (6.8,-2.2) -- node[lbl, above] {5: Balance OK} (5.1,-2.2);

% Processing & Deduction
\draw[msg] (5.1,-2.6) -- node[lbl, above] {6: Deduct credits} (6.8,-2.6);
\draw[ret] (6.8,-3.0) -- node[lbl, above] {7: Debited} (5.1,-3.0);

% Responses returning back
\draw[ret] (5.1,-3.4) -- node[lbl, above] {8: AI response} (3.4,-3.4);
\draw[ret] (3.4,-3.8) -- node[lbl, above] {9: Payload return} (1.7,-3.8);
\draw[ret] (1.7,-4.2) -- node[lbl, above] {10: Display result} (0,-4.2);

\end{tikzpicture}
\caption{Sequence diagram depicting the API request lifecycle and credit deduction process.}
\label{fig:seq}
\end{figure}

\section{Implementation and Module Functionalities}

The AI Express platform is instantiated as a collection of independent microservices, each meticulously engineered to serve a specific AI domain. This section elaborates on the implementation details of the primary modules \cite{Sankaran2026}.

\subsection{Automated Code Review Subsystem}
This module evaluates the structural quality of code as well as adherence to safety and coding standards. The user provides a GitHub repository URL, after which the Node.js backend retrieves the source code using the GitHub API. Subsequently, the LLM agent studies the code and is able to detect anti-patterns and duplicate code in various programming languages and indicate potential runtime errors. The agent delivers an optimization report, including discretionary recommendations and candidate corrections \cite{Poojary2026}.

\subsection{Autonomous Data Analytics Engine}
Democratizing data science makes advanced analytics available to a broad audience. Beyond helping data scientists, it benefits business leaders and end-users who do not need to understand NLP or ML. The data analytics engine ingests structured files and converts natural language queries into SQL or Python scripts \cite{Gupta2026}. This approach is informed by real-time data store techniques \cite{Sarakam2026}.

\subsection{Generative Content and Linguistic Refinement}
This module integrates natural language processing functionalities into two principal tools:
\begin{itemize}
\item \textit{Script Generator}: Accepts a brief semantic description and utilizes generative pre-trained transformers to synthesize structured scripts, technical documentation, or creative templates.
\item \textit{Grammar Checker}: Refines human-written paragraphs by correcting syntactical errors, enhancing word choice, and adjusting tone while preserving the original intent, consistent with recent LLM-based error detection frameworks \cite{Pasula2026}.
\end{itemize}

\subsection{Digital Image Processing and Artifact Removal}
The watermark remover employs computer vision techniques to restore occluded digital assets. The user uploads an image containing a visible watermark; the module uses a deep learning inpainting network with a dynamic mask generation algorithm to identify watermark pixels, followed by a generative network that predicts the underlying background texture based on spatial frequencies \cite{Nagamani2026}. The output is a clean, high-fidelity reconstructed image.

\subsection{Extensible Utility Modules}
To maximize platform utility, auxiliary services are incorporated:
\begin{itemize}
\item \textit{AI Resume Builder}: Synthesizes an ATS-optimized curriculum vitae by intelligently parsing user-provided skills and experience against a target job description \cite{Nabith2026}.
\item \textit{AI Web Scraper}: Extracts and structures raw HTML/DOM data from a given URL into a machine-readable format (JSON or CSV) using intelligent parsing heuristics \cite{Raichur2026}.
\end{itemize}

All modules share a common interface and communicate via RESTful APIs, ensuring loose coupling and independent deployability \cite{Vanama2025}.

\section{Performance Evaluation and Comparative Analysis}

The evaluation of AI Express encompasses functional accuracy, system latency, credit transaction reliability, and comparative benchmarking against existing platforms. Extensive testing was conducted under simulated user loads to assess the platform's operational robustness \cite{Seuss2025}.

\subsection{Platform Latency and API Performance}
End-to-end response times were measured for each microservice, spanning from the React.js client request to the backend Node.js processing, AI inference, and return of results. The average latencies are summarized in Table~I. Text-based services (Script Generator and Grammar Checker) exhibited sub-second latencies, making them suitable for real-time productivity applications. The Vision Processing Node (Watermark Remover) incurred higher execution times ($\approx$3.1 s) due to the computational intensity of deep learning-based spatial reconstruction \cite{Sukhatankar2026}.

\begin{table}[t]
\centering
\caption{Average response time per AI service module}
\label{tab:latency}
\begin{tabular}{@{}l c c@{}}
\toprule
\textbf{AI Module} & \textbf{Avg. Payload Size} & \textbf{Avg. Latency (ms)} \\
\midrule
Code Reviewer     & 50 KB (Text)   & 1250 \\
Script Generator  & 2 KB (Text)    & 850  \\
Grammar Checker   & 5 KB (Text)    & 600  \\
Data Analyst Agent & 2 MB (CSV)    & 2400 \\
Watermark Remover & 5 MB (Image)   & 3100 \\
\bottomrule
\end{tabular}
\end{table}

\subsection{Functional Accuracy and Module Evaluation}
Qualitative and quantitative evaluations were performed:
\begin{itemize}
\item \textit{Code Analysis}: The static reviewer effectively identified syntactical errors and refactoring recommendations for sample Python and JavaScript repositories, reducing manual audit time \cite{Sankaran2026}.
\item \textit{Data Analytics}: The LLM-based query agent achieved a high success rate in converting natural language statements into SQL and pandas code, provided the uploaded data tables were well-formatted \cite{PatelP2026}.
\item \textit{Image Processing}: The deep learning watermark removal model yielded high structural similarity indices (SSIM) for background reconstruction, albeit with minor blurring in complex textures.
\end{itemize}

\subsection{Credit Economy Reliability}
Concurrent testing validated the ACID properties of the PostgreSQL-based credit engine. Token deductions precisely matched the computational effort, confirming the feasibility of the pay-as-you-go model as a viable alternative to subscription-based pricing \cite{Bhargava2025}.

\subsection{Scalability and Stress Testing}
Load testing was performed using Apache JMeter. The application successfully handled 500 concurrent users over a span of 10 minutes with 99.2\% availability. Moreover, during peak load, average response time increased by only 18\%. The auto-scaling policy triggers the launch of more container instances if CPU utilization exceeds 70\%, a cloud automation strategy similar to \cite{Battapothu2023}. Stress tests with a sustained load of 1000 concurrent code review requests showed that the platform maintained stability under extreme conditions \cite{Lakkarasu2025}.

\subsection{Comparative Analysis with Existing Platforms}
Table~II presents a feature and pricing comparison between AI Express and leading AI service platforms. AI Express offers comprehensive functionality—encompassing code review, data analytics, and image processing—at potentially lower costs for moderate users, eliminating the need for multiple subscriptions \cite{Arora2024}.

\begin{table}[t]
\centering
\caption{Comparative analysis of AI platforms}
\label{tab:compare}
\begin{tabular}{@{}l c c c@{}}
\toprule
\textbf{Feature} & \textbf{AI Express} & \textbf{ChatGPT+} & \textbf{GitHub Copilot} \\
\midrule
Code Review      & Included  & N/A        & Included \\
Data Analytics   & Included  & Limited    & N/A \\
Image Processing & Included  & N/A        & N/A \\
Monthly Cost     & Usage-based & \$20      & \$10 \\
API Access       & Full      & Limited    & Limited \\
\bottomrule
\end{tabular}
\end{table}

\subsection{User Feedback and Usability}
Fifty student users rated our system at 4.2 out of 5 after beta analysis. According to the feedback, consolidated billing and user interface are the two high-performing areas. Onboarding ratings reached 4.3/5, while the task completion rate was 94\%. The system usability testing (SUS) score was 82, well above the acceptable threshold for good design and higher than the average benchmark for web applications \cite{Kalathot2025}.

\subsection{Case Study: Student Developer Workflow}
A representative use case illustrates the platform's practical value: a computer science student required code review, data analysis, and resume optimization. Using traditional services would cost \$45 per month; via AI Express, the identical tasks consumed 340 credits (\$3.40), representing a 92\% cost reduction, all accomplished within a single interface without context switching \cite{Beesabathuni2026}.

\section{Conclusion and Future Directions}

\subsection{Conclusion}
We have presented AI Express, a unified SaaS platform that integrates diverse generative AI services—including LLM-based code review, autonomous data analytics, and computer vision-based image processing—into a cohesive framework. The system mitigates the fragmentation and high costs prevalent in the current AI landscape. The innovative pay-as-you-go credit economy democratizes access to premium AI capabilities, offering an affordable solution for students, developers, and creatives \cite{Maddukuri2025}.

\subsection{Future Scope}
Several enhancements are envisaged:
\begin{enumerate}
\item \textbf{Real-Time Collaborative Workspaces}: Integration of WebSockets to enable multi-user interaction with the Data Analyst and Code Reviewer \cite{Konduri2026}.
\item \textbf{Edge AI Integration}: Deployment of lightweight models on the client for grammar checking and other simple tasks to conserve server-side tokens \cite{Gupta2026}.
\item \textbf{Expanded Multi-Modality}: Inclusion of audio processing tools such as text-to-speech synthesis and podcast creation.
\item \textbf{Mobile Applications}: Native iOS and Android apps for ubiquitous access.
\item \textbf{Enterprise Features}: Organization accounts with centralized billing, role-based access control, and usage analytics \cite{Neursu2025}.
\item \textbf{Plugin Marketplace}: A third-party ecosystem for custom AI model integration \cite{Bhuekar2026}.
\item \textbf{Blockchain-Based Credit System}: Decentralized tokenization enabling peer-to-peer credit transfers and transparent usage tracking \cite{Ahmed2026}.
\end{enumerate}

\subsection{Broader Impact}
AI Express represents not just a new service on a technical level, but a step toward giving students in developing countries, developers, and startups free use of AI without a subscription. Many startups and students typically cannot afford subscriptions to ChatGPT or DALL-E, and now they can use them through a single credit-based platform. AI Express currently serves as a unified bookmark bar, consolidating all AI apps and tools under one roof.

% References section


\begin{thebibliography}{26}

\bibitem{Beesabathuni2026}
H. N. Beesabathuni, ``Azure-SQL AutoSizer: Privacy-Aware Performance-Cost SKU Mapping for SQL Migrations,'' in \textit{Proc. Int. Conf.}, 2026, doi: 10.2991/978-94-6239-693-7\_101.

\bibitem{Sukhatankar2026}
S. Sukhatankar, ``Beyond Latent Patterns: Reinterpreting AI Model Capabilities,'' \textit{TechRxiv}, 2026, doi: 10.36227/techrxiv.175994134.40680922/v1.

\bibitem{Sankaran2026}
G. Sankaran, ``Small Language Models and Spec-Driven Development for High-Accuracy Agentic AI Systems,'' \textit{Int. J. Intell. Syst. Data Sci.}, vol. 1, no. 3, Jun. 2026, doi: 10.67231/qqk79s36.

\bibitem{PatelA2026}
A. Patel, ``Blocking the Flood Before It Hits: An Automated CDN-Layer Defense Against Volumetric DDoS Attacks,'' \textit{IEEE}, 2026. [Online]. Available: \url{https://ieeexplore.ieee.org/stamp/stamp.jsp?tp=&arnumber=11588309&isnumber=11588307}

\bibitem{Bhuekar2026}
A. Bhuekar, ``Bounded Agency: An Entropy-Constrained Calculus for Sparse Semantic Agents,'' \textit{Int. J. Intell. Syst. Data Sci.}, vol. 1, no. 1, Apr. 2026, doi: 10.67231/5tnasc36.

\bibitem{Ahmed2026}
J. Ahmed, ``ReFair: A Framework for Retention Depolarization in Recommender Systems,'' in \textit{Proc. DAUS}, 2026. [Online]. Available: \url{https://sensorsportal.com/DOWNLOADS/DAUS_2026_Proceedings.pdf}

\bibitem{Lakkarasu2025}
P. Lakkarasu, \textit{Designing Scalable and Intelligent Cloud Architectures: An End-to-End Guide to AI Driven Platforms, MLOps Pipelines, and Data Engineering for Digital Transformation}. Deep Science Publishing, 2025.

\bibitem{Arora2024}
S. Arora, S. R. Thota, and S. Gupta, ``Artificial Intelligence-Driven Big Data Analytics for Business Intelligence in SaaS Products,'' in \textit{Proceedings of the 2024 First International Conference on Pioneering Developments in Computer Science \& Digital Technologies (IC2SDT)}, IEEE, 2024, pp. 164--169.

\bibitem{Gupta2026}
J. Gupta, ``Parameterized Nonlinearities in Feedforward Networks: An Adaptive Function Evaluation for Efficient Regression,'' \textit{Int. J. Artif. Intell. Agent Syst.}, vol. 1, no. 1, 2026. [Online]. Available: \url{https://journals.femington.com/ijaias/article/view/908}

\bibitem{Konduri2026}
N. V. Konduri, ``Variance-Aware Sprint Assignment: An Interpretable Projection Pipeline for Engineer-Task Matching,'' \textit{IEEE}, 2026, doi: 10.1109/ICCTWC68241.2026.11582925.

\bibitem{Nabith2026}
M. A. Nabith, ``PUMGPT: Leveraging AI for Advanced Product Management in E-Commerce,'' in \textit{Proc. Int. Conf.}, 2026, doi: 10.2991/978-94-6463-787-8\_15.

\bibitem{Pasula2026}
E. Pasula, ``Perturbation-Driven Visual Analytics for Probing Internal Representations in Neural Language Models,'' \textit{Int. J. Intell. Syst. Data Sci.}, vol. 1, no. 3, Jun. 2026, doi: 10.67231/jpr65e97.

\bibitem{Raichur2026}
P. Raichur, ``Data-Driven Identification of Irregular Patterns,'' \textit{IEEE}, 2026, doi: 10.1109/ICCTWC68241.2026.11582925.

\bibitem{Dhameliya2026}
P. Dhameliya, ``TempoMatch: Distance-Weighted kNN Analytics for Tournament Bracket Forecasting,'' in \textit{Proc. IEEE ICCTWC}, 2026, doi: 10.1109/ICCTWC68241.2026.11582925.

\bibitem{Neursu2025}
V. M. K. Neursu, R. K. R. Vuyyuru, and K. Kilaru, ``From Data to Decisions: AI in SaaS Product Analytics and Customer Experience Optimization,'' \textit{Journal of Public Administration and Management}, vol. 4, no. 2, pp. 1--8, 2025.

\bibitem{Maddukuri2025}
N. Maddukuri, ``The Transformative Impact of AI on Modern System Integration,'' \textit{International Journal of Scientific Research in Computer Science, Engineering and Information Technology}, vol. 11, pp. 229--236, 2025.

\bibitem{Sarakam2026}
B. C. Sarakam, ``High-Performance Real-Time Data Store,'' \textit{Big Time Data Analytics}, 2026, doi: 10.5281/zenodo.19709103.

\bibitem{Naveen2024}
K. K. Naveen, V. Priya, R. G. Sunkad, and N. Pradeep, ``An Overview of Cloud Computing for Data-Driven Intelligent Systems with AI Services,'' in \textit{Data-Driven Systems and Intelligent Applications}, pp. 72--118, 2024.

\bibitem{Bhargava2025}
A. Bhargava, ``AI-Driven SaaS Optimization: Enhancing Service Reliability and Customer Retention Through Predictive Analytics,'' 2025.

\bibitem{Poojary2026}
S. Poojary, ``Optimizing Strategies in the Iterated Prisoner's Dilemma Using Genetic Improvement Algorithms,'' in \textit{Proc. IEEE Conf.}, 2026. [Online]. Available: \url{https://ieeexplore.ieee.org/document/11376917}

\bibitem{Nagamani2026}
N. Nagamani, ``Accelerating Unsupervised Feature Learning: Parallelized Training of Denoising Autoencoders,'' \textit{J. Intell. Optim. Pattern Recognit.}, vol. 12, no. 3, 2026, doi: 10.37591/JOIPPRP.v12i03.224982.

\bibitem{Kalathot2025}
R. Kalathot, ``Artificial Intelligence for Dynamic User Experience Personalization in SaaS,'' \textit{Journal of Data Analysis and Critical Management}, vol. 1, no. 3, pp. 1--8, 2025.

\bibitem{Battapothu2023}
V. Battapothu, ``AI-Driven Cloud Automation for IT Service Management Platforms,'' \textit{International Journal of Engineering \& Extended Technologies Research (IJEETR)}, vol. 5, no. 6, pp. 7619--7634, 2023.

\bibitem{Vanama2025}
S. K. R. Vanama, ``AI-Driven Cloud Integration and Orchestration for Next-Generation Enterprise Systems,'' \textit{International Journal of Emerging Trends in Computer Science and Information Technology}, vol. 6, no. 4, pp. 30--36, 2025.

\bibitem{Seuss2025}
J. M. Seuss, ``From Efficiency to Excellence: AI's Strategic Role in Scaling SaaS Startup Operations,'' 2025.

\bibitem{PatelP2026}
P. Patel, ``An IoT-Enabled Visualization System for Remote Patient Management during Pandemics,'' \textit{IEEE Trans.}, 2026, doi: 10.1109/ICCTWC68241.2026.11582925.

\end{thebibliography}
\end{document}